\documentclass[11pt]{article}

\usepackage{paperstyle}

\title{Target-Aligned Bellman Backup for \\
Cross-domain Offline Reinforcement Learning}

\author{
\textbf{Wei Liu}\\
School of Artificial Intelligence\\
Jilin University\\
China\\
\texttt{lwei25@mails.jlu.edu.cn}
\And
\textbf{Ting Long}\textsuperscript{*}\\
School of Artificial Intelligence\\
Jilin University\\
China\\
\texttt{longting@jlu.edu.cn}
}

\date{}

\begin{document}

\maketitle

\begin{abstract}
Cross-domain offline reinforcement learning (CDRL) aims to improve policy learning in a target domain by leveraging data collected from a source domain.
Existing works typically assess the transferability of source-domain data by measuring its similarity to target-domain transitions, and implicitly perform transition-level selection. Transitions that are considered similar are assigned higher weights or rewards, while dissimilar ones are down-weighted.
However, transition-level similarity does not necessarily imply consistency in long-term returns. Even visually or dynamically similar transitions may lead to significantly different outcomes in the target domain, which can mislead policy learning and degrade performance.
To address this issue, we revisit the fundamental objective of policy learning. Since policy optimization ultimately relies on Bellman targets to evaluate the quality of decisions, we propose to assess the transferability of source-domain transitions based on their alignment with target-domain Bellman targets, rather than superficial transition similarity. Based on this insight, we propose a method termed Target-Aligned Bellman Backup (TABB), which selectively leverages source-domain data by measuring their contribution to accurate Bellman target estimation in the target domain.
We evaluate TABB across a broad range of cross-domain offline RL settings with highly limited target-domain data. Experimental results show that TABB consistently achieves strong performance.
\end{abstract}

\section{Introduction}

Cross-domain offline reinforcement learning (CDRL) is a promising paradigm that leverages source-domain data to improve target-domain policy learning~\cite{wen2024contrastive,liu2024beyond,lyu2025cross}. By reusing existing datasets across domains, it fundamentally enhances data efficiency and reduces reliance on costly target-environment interactions. This paradigm is particularly valuable in real-world decision making scenarios, such as robotics \cite{levine2020offline,ebert2021bridge}, recommendation systems \cite{zhang2022text,xue2022prefrec,xiao2021general}, and autonomous control \cite{kidambi2020morel,xu2022constraints,kiran2021deep}, where direct interaction with the target environment is limited or expensive.

\begin{wrapfigure}{r}{0.5\linewidth}
    \vspace{-15pt}
    \centering
    \includegraphics[width=\linewidth]{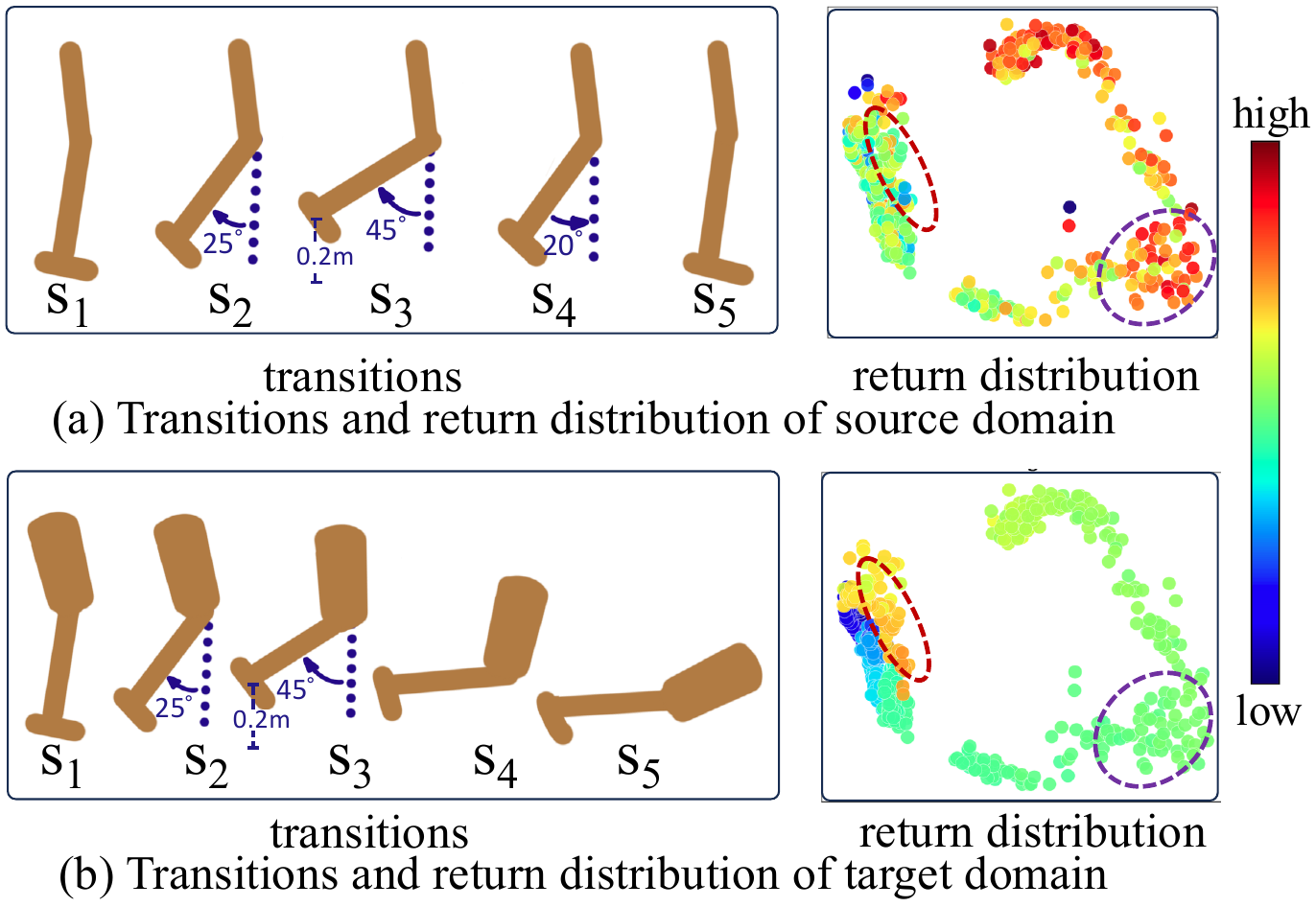}
    \vspace{-15pt}
    \caption{Transition and return distribution}
    \label{fig:intro}
\vspace{-15pt}
\end{wrapfigure}

Despite its appeal, directly applying source-domain data to the target domain often suffers from distribution shifts and task mismatches, which can lead to suboptimal or even detrimental policy learning. 
To mitigate this issue, prior work~\cite{eysenbach2020off,liu2022dara,xu2023cross,liu2024beyond,wen2024contrastive,lyu2025cross} typically assesses the utilization of source-domain data by estimating its similarity to target-domain transitions, and implicitly performs transition-level selection based on this criterion. Specifically, they either assign higher rewards to transitions that better match the target domain~\cite{liu2022dara}, or reweight them during training to amplify their influence while suppressing less relevant ones~\cite{wen2024contrastive,lyu2025cross}.
As illustrated in Fig.~\ref{fig:intro}, transitions $(\textbf{s}_2, \textbf{a}_2, \textbf{s}_3)$ in source domain and transition $(\textbf{s}_2, \textbf{a}_2, \textbf{s}_3)$ in target domain appear highly similar, as both agents gradually lifts its leg to a height of 0.2 meters with an angle of $45^\circ$ relative to the horizontal. Consequently, existing methods would assign relatively high importance (e.g., rewards or weights) to these source transitions and incorporate the reweighted source data together with the original target data for policy learning.

However, such strategies can be problematic.
As illustrated in Fig.~\ref{fig:intro}, although the transition $(\textbf{s}_2, \textbf{a}_2, \textbf{s}_3)$ in the source domain appears highly similar to $(\textbf{s}_2, \textbf{a}_2, \textbf{s}_3)$ in the target domain, the resulting returns can differ significantly. In the source domain, due to the lighter upper body, executing action $\textbf{a}_2$ at $\textbf{s}_2$ (e.g., lifting the leg to form a $45^\circ$ angle with the horizontal) helps the agent achieve a longer jump and land stably at step 5. In contrast, in the target domain, where the upper body is heavier, executing the same action at $\textbf{s}_2$ may cause instability, leading the agent to fall forward and fail (state $\textbf{s}_5$).
In fact, it is common for similar transitions to have substantially different returns.
To further investigate that, we visualize the returns of matched transitions from the source and target domains in two separate plots, as shown in Fig.~\ref{fig:intro}. Each scatter represents a transition projected into a shared low-dimensional space, where corresponding regions across the two plots denote similar transitions. The color of each scatter encodes the return associated with the initial state of that transition.
We observe that, even within corresponding regions, the return distributions can differ substantially between the source and target domains:
regions yielding high returns in the source domain may correspond to low-return regions in the target domain, and vice versa, as highlighted by the dashed circles.
This suggests that reweighting or sampling based solely on transition similarity may introduce bias and noise into target-domain Bellman updates. Since policy learning relies on iterative Bellman updates~\cite{fujimoto2019off,kumar2019stabilizing,kumar2020conservative}, such inconsistencies can lead to inaccurate value propagation and ultimately degrade policy performance.

Motivated by the above observations, we revisit cross-domain offline RL from the perspective of transition similarity to the core principle of policy learning: the Bellman backup. We propose a method termed \textbf{Target-Aligned Bellman Backup (TABB)}, which selects and reweights source-domain data based on their alignment with Bellman backup consistency in the target domain, rather than relying on transition-level similarity.
Specifically, TABB assigns higher weights to source transitions whose induced Bellman backups are consistent with those in the target domain, and down-weights mismatched ones. The reweighted source data is then combined with the target data to train the target policy.

We conduct extensive experiments across 6 environments, 16 dataset combinations, covering two types of cross-domain transfer settings. 
The results demonstrate that TABB consistently achieves strong performance improvements over existing methods.

In summary, our work has the following contributions:
\begin{itemize}  [leftmargin=1.5em]

\item We identify that transition-level similarity does not guarantee consistency in value updates in cross-domain offline RL, which can introduce bias in policy learning.
\item We propose Target-Aligned Bellman Backup (TABB), which selects and reweights source data based on its Bellman backup consistency with the target domain, rather than transition similarity.
\item We demonstrate that TABB consistently improves the performance across multiple environments of cross-domain offline RL settings.
\end{itemize}

\section{Related Work}

Cross-domain offline RL aims to mitigate data scarcity in a target domain by leveraging the data in the source domain, which is collected under different transition dynamics~\citep{van2025dmc, nishimori2024offline, mark2024policy, kumar2022pre}. A central challenge is to exploit source data without inducing negative transfer under dynamics mismatch. To handle this issue, prior methods can be broadly categorized into regularization-based methods~\cite{liu2022dara,wang2024return, liu2024beyond}, filtering-based methods~\cite{wen2024contrastive,lyu2025cross}, and sample-generation-based methods~\cite{van2025dmc,guo2025mobody}.
Despite differences in implementation, these methods share a common principle: they estimate the transferability of source data primarily based on source--target transition similarity, and determine how source transitions are incorporated into target policy learning.

\textit{Regularization-based methods} address dynamics mismatch by incorporating source--target transition similarity into the learning objective through additional regularization terms or constraints. 
For example, DARA~\cite{liu2022dara} estimates transition-level dynamics mismatch and augments source rewards accordingly. 
REAG~\cite{wang2024return} extends this augmentation principle to return-conditioned supervised learning by transforming source returns to better reflect their target-domain compatibility. 
In contrast, BOSA~\cite{liu2024beyond} handles mismatch through supported policy optimization and supported value optimization, explicitly accounting for both out-of-distribution state-action pairs and out-of-distribution transition dynamics. 
SRPO~\cite{xue2023state} approaches the problem from a different perspective by regularizing the learned policy toward a reference stationary state distribution across related dynamics, thereby constraining policy learning within transition distributions that are more compatible with the target domain.

\textit{Filtering-based methods} address dynamics mismatch by explicitly using source--target transition similarity as a criterion for selecting source-domain transitions. 
A representative example is IGDF~\cite{wen2024contrastive}, which learns contrastive transition representations to measure the cross-domain gap and filters source transitions accordingly during value learning. 
OTDF~\cite{lyu2025cross} further performs transition-level alignment through optimal transport and filters source samples based on the resulting alignment scores, enabling more reliable source-data selection, especially when target-domain supervision is limited.

\textit{Sample-generation-based methods} address dynamics mismatch by using source--target transition similarity to generate additional target-compatible training samples beyond the original offline datasets. 
DmC~\cite{van2025dmc}uses a nearest-neighbor-guided diffusion model to generate additional source samples that are better aligned with the target transition distribution. 
MOBODY~\cite{guo2025mobody}, in contrast, learns a target-domain dynamics model from source and limited target data, and then performs model rollouts to synthesize target-domain transitions for target policy optimization.

\section{Preliminaries}

\textbf{Cross-Domain Offline RL.} 
We consider the standard cross-domain offline RL setting, where the target environment is modeled as a Markov Decision Process (MDP) $\mathcal{M}_{\mathrm{tar}} = (\mathcal{S}, \mathcal{A}, P_{\mathrm{tar}}, r, \gamma)$, with state space $\mathcal{S}$, action space $\mathcal{A}$, transition dynamics $P_{\mathrm{tar}}(\textbf{s}' \mid \textbf{s},\textbf{a})$, reward function $r(\textbf{s},\textbf{a})$, and discount factor $\gamma \in [0,1)$~\cite{bellman1957markovian,puterman1990markov,kaelbling1996reinforcement}. Unlike online reinforcement learning, offline RL access only to a fixed target-domain dataset 
$\mathcal{D}_{\mathrm{tar}}$ consisting of transitions $(\textbf{s},\textbf{a},r,\textbf{s}')$, without any additional interaction with the environment~\cite{levine2020offline,fujimoto2019off}. Besides, we assume access to an offline dataset from a source domain $\mathcal{D}_{\mathrm{src}}$ consist of transition $(\textbf{s},\textbf{a},r,\textbf{s}')$, which is collected from the source-domain MDP $\mathcal{M}_{\mathrm{src}}=(\mathcal{S},\mathcal{A},P_{\mathrm{src}},r,\gamma)$. As in prior cross-domain offline RL works~\cite{lyu2024odrl,lyu2025cross}, we assume that the source and target domains share the same state space and action space, and differ only in their transition dynamics ($P_{\mathrm{src}} \neq P_{\mathrm{tar}}$). In addition, the target-domain dataset is typically much smaller than the source-domain dataset, with $|\mathcal{D}_{\mathrm{tar}}| \ll |\mathcal{D}_{\mathrm{src}}|$. 
Our goal is to leverage the mixed dataset $\mathcal{D}_{\mathrm{mix}} = \mathcal{D}_{\mathrm{src}} \cup \mathcal{D}_{\mathrm{tar}}$ to learn a policy $\pi$ that maximizes the cumulative reward under the target dynamics:
\begin{equation} \label{eq:learning_target}
\pi^*=
\arg\max_{\pi}\;
\mathbb{E}_{\tau \sim P_{\mathrm{tar}}, \pi}
\left[
\sum_{t=0}^{\infty} \gamma^t r(\textbf{s}_t,\textbf{a}_t)
\right], 
\end{equation}
where $t$ indexes time steps.

\textbf{Bellman backup and Bellman target.} 
Achieving Eq.~\eqref{eq:learning_target} typically requires evaluating the quality of different decisions, which is commonly done through value functions that estimate the expected return (\textit{i.e.}, value) of states. However, directly estimating the value of states is challenging due to their dependence on future, unknown outcomes.
To address this, the Bellman backup~\cite{bellman1966dynamic,sutton1998reinforcement} provides a recursive decomposition of the value function, expressing the value of a state as the combination of its immediate reward and the value of subsequent states. Given a transition $\xi=(\textbf{s},\textbf{a},r,\textbf{s}')$, the Bellman backup can be written as:
\begin{equation}
V(\textbf{s}) \leftarrow \mathbb{E} \big[ r + \gamma V(\textbf{s}') \big].
\end{equation}
The right-hand side is referred to as the Bellman target: $y(\xi)=r+\gamma V(\textbf{s}')$. In practice, the value function is learned by fitting to this target, typically by minimizing the Bellman error between the predicted value and the target:
\begin{equation}
\mathcal{L}(\theta)=\mathbb{E}_{(\textbf{s},\textbf{a},r,\textbf{s}')\sim \mathcal{D}}\left[\big(V_\theta(s)-y(\xi)\big)^2\right].
\end{equation}
The learned value function $V_\theta(s)$ subsequently guides policy optimization, either by directly informing action selection or by providing training signals for policy improvement.
This highlights that the quality of the learned policy critically depends on the accuracy of the value function, which in turn is determined by the quality of the Bellman target.
When source-domain data exhibit significant discrepancies from the target domain in terms of Bellman targets, incorporating such data can introduce bias into target value estimation, thereby degrading policy performance.
Accordingly, the transferability of a source transition is better characterized by the consistency of its Bellman target with that of the target domain.

\section{Method}

Motivated by the above analysis, we revisit cross-domain offline RL from the perspective of the Bellman target, the key factor governing policy performance, and propose a method called \emph{Target-Aligned Bellman Backup} (TABB). It first introduces the \emph{Target Bellman Mismatch} (TBM) to quantify the deviation of a source transition under the target-domain Bellman backup. Based on TBM, it softly reweights source transitions according to their transferability, and integrates the weighted source data with target data to train the target policy. In the following, we first present the details of TBM, and then describe how to reweight source data with TBM and leverage the reweighted source data for policy training.

\subsection{Target Bellman Mismatch.}
\label{sec:tbm}
The \emph{Target Bellman Mismatch} (TBM) is designed to measure the discrepancy between the Bellman target of a source transition under the source domain and its Bellman backup under the target domain. 
Intuitively, a small discrepancy suggests that the source transition can provide reliable supervision for the target-domain Bellman target, whereas a large discrepancy indicates that it may induce bias to Bellman target and thus increase the risk of negative transfer. 

Formally, given a source transition $\xi=(\mathbf{s},\mathbf{a},r,\mathbf{s}')$, we first encode the state and state-action pair into a shared latent space as:
\begin{equation}
\label{eq:latent_rep}
\mathbf{z}_{s}=\phi(\mathbf{s}),\qquad 
\mathbf{z}_{sa}=\psi(\mathbf{s},\mathbf{a}),\qquad
\mathbf{z}'_{s}=\phi(\mathbf{s}').
\end{equation}
Here, $\phi$ denotes the state encoder, and $\psi$ denotes the state-action encoder. $\mathbf{z}_{s}$ and $\mathbf{z}_{sa}$ denote the state and state-action representations in latent space, respectively.

Next, we construct two Bellman targets for $\xi$: a realized Bellman target 
$y_{\mathrm{real}}(\xi)$, computed from its actual reward and next state, and a target-aligned Bellman target $y_{\mathrm{tar}}(\xi)$, 
computed from the predicted reward and next-state by a target-domain predictor $f_{\mathrm{ref}}$, and define the TBM as the absolute difference between these two targets:  
\begin{equation}
\label{eq:tbm}
\mathrm{TBM}(\xi)
=
\left|
\underbrace{\left(r+\gamma V(\mathbf{z}'_{s})\right)}_{y_{\mathrm{real}}(\xi)}
-
\underbrace{\left(\hat r+\gamma V(\hat{\mathbf{z}}'_{s})\right)}_{y_{\mathrm{tar}}(\xi)}
\right|.
\end{equation}
Here, $(\hat r,\hat{\mathbf{z}}'_s)=f_{\mathrm{ref}}(\mathbf{z}_{s},\mathbf{z}_{sa})$. $f_{\mathrm{ref}}$ denotes the target-domain predictor implemented by a multi-layer perceptron (MLP). 
$V$ denotes a target-domain anchor value function in latent space, which is learned exclusively from target data. We use the same $V$ in computing both $y_{\mathrm{real}}(\xi)$ and $y_{\mathrm{tar}}(\xi)$ so that TBM isolates the discrepancy caused by the reward and next-state mismatch of the same source state-action pair. Otherwise, the difference would also reflect the mismatch between different value functions themselves. Moreover, because $V$ is trained only on target data, TBM remains anchored to target-domain Bellman updates. 

We learn the encoders $\phi$, $\psi$, and the target-domain predictor $f_{\mathrm{ref}}$ in a two-stage strategy.

In the first stage, we leverage both the source and target data to jointly train the encoders $\phi$, $\psi$, and the target-domain predictor $f_{\mathrm{ref}}$:
\begin{equation} \label{eq:tbm_loss}
\mathcal{L}_{\mathrm{tmb}}
=
\mathbb{E}_{(\textbf{s},\textbf{a},r,\textbf{s}')\sim\mathcal{D}_{\mathrm{src}} \cup \mathcal{D}_{\mathrm{tar}}}
\left[
(\hat r-r)^2 +
\mathbf{W}_{\mathrm{tar}}
\left\|
\mathbf{z'}_{s} -\hat{\mathbf{z'}}_{s}
\right\|_2^2
\right] .
\end{equation}
Here, $\mathbf{W}_{\mathrm{tar}}$ is a diagonal matrix computed from $\textbf{Z}_s^\text{tar}$, each row of which represents a latent representation ($\mathbf{z}_{s}$) of a target state $\textbf{s}$. Specifically, each diagonal element is determined by the variance of the corresponding latent dimension in $\mathbf{z}_s$ over target-domain transitions.

Since our ultimate goal is to learn a policy for the target domain, the predictor $f_{\mathrm{ref}}$ obtained from the Eq.~\eqref{eq:tbm_loss} may still be affected by source-domain dynamics. Therefore, in the second stage, we freeze the parameters of the encoders $\phi$, $\psi$, and further fine-tune $f_{\mathrm{ref}}$ with respect to target-domain dynamics by replacing $\mathcal{D}_{\mathrm{src}} \cup \mathcal{D}_{\mathrm{tar}}$ in Eq.~\eqref{eq:tbm_loss} with $ \mathcal{D}_{\mathrm{tar}}$.
It is worth noting that we do not train the model using only target-domain data in the first stage, as in real-world scenarios the amount of target-domain data is typically limited, making it insufficient to learn a high-quality encoder.

\subsection{Reweighting Source Data for Policy Learning}
Building on the TBM defined in Section~\ref{sec:tbm}, we reweight the transitions of the source domain, and we leverage the weighted source data with the target data to train the target policy. 
Specifically, for a source transition $\xi=(\textbf{s},\textbf{a},r,\textbf{s}')\in\mathcal{D}_{\mathrm{src}}$, we convert its TBM into the transferability weight:
\begin{equation} \label{eq:weight}
\omega(\xi)
=
\frac{
\exp\!\left(-\mathrm{TBM}(\xi)\right)
}{
\sum_{\xi'\in\mathcal{B}_{\mathrm{src}}}
\exp\!\left(-\mathrm{TBM}(\xi')\right)
},
\end{equation}
where $\mathcal{B}_{\mathrm{src}}$ denotes the source transitions in batch. 
Transitions with smaller TBM values receive larger weights, since their Bellman targets are more consistent with the target-aligned targets. We then update the critic using both target transitions and weighted source transitions. For both source and target data, the Bellman target follows the IQL-style update $y_{\mathrm{tar}}(\xi)=r+\gamma V(\phi(\textbf{s}'))$.  
The critic objective is:
\begin{equation}
\label{eq:q_loss}    
\mathcal{L}_{Q}
=
\underbrace{
\mathbb{E}_{\xi\sim\mathcal{D}_{\mathrm{tar}}}
\left[
\ell_{\mathrm{td}}\big(q_\theta(\xi),y_{\mathrm{tar}}(\xi)\big)
\right]
}_{\mathcal{L}_{Q}^{\mathrm{tar}}}
+
\underbrace{
\mathbb{E}_{\xi\sim\mathcal{D}_{\mathrm{src}}}
\left[
\omega(\xi)\,
\ell_{\mathrm{td}}\big(q_\theta(\xi),y_{\mathrm{tar}}(\xi)\big)
\right]
}_{\mathcal{L}_{Q}^{\mathrm{src}}}.
\end{equation}
Here, $q_\theta(\xi)=Q_\theta(\phi(\textbf{s}),\psi(\textbf{s},\textbf{a}))$ denotes the critic prediction in the shared latent space. We adopt the state-action representation $\psi(\textbf{s},\textbf{a})$ for Q-value estimation, as actions are not explicitly encoded in the latent space. The representation $\psi(\textbf{s},\textbf{a})$ implicitly captures action information, enabling accurate value prediction. 
$\ell_{\mathrm{td}}$ is the critic regression loss, implemented with the Huber loss for robustness to outliers in Bellman target regression. The transition weight $\omega(\xi)$ controls the relative reliability of each source sample.
Source data update $Q_\theta$ through the target-aligned Bellman target $y_{\mathrm{tar}}(\xi)$, avoiding direct backups from source dynamics that may be inconsistent with target-domain value learning.

The value function and policy are updated with target-domain data following standard IQL updates. Specifically, $V$ is fitted by the expectile objective:
\begin{equation}
\label{eq:v_loss} 
\mathcal{L}_{V}
=
\mathbb{E}_{\xi\sim\mathcal{D}_{\mathrm{tar}}}
\left[
L_2^{\tau}\big(q_\theta(\xi)-V(\phi(\textbf{s}))\big)
\right],    
\end{equation}
where $L_2^{\tau}$ denotes the asymmetric squared loss. The policy is extracted by advantage-weighted regression:
\begin{equation}
\label{eq:pi_loss} 
\mathcal{L}_{\pi}
=
\mathbb{E}_{\xi\sim\mathcal{D}_{\mathrm{tar}}}
\left[
\exp\!\left(
\beta\big(q_\theta(\xi)-V(\phi(\textbf{s}))\big)
\right)
\|\pi(\textbf{s})-\textbf{a}\|_2^2
\right].
\end{equation}
Thus, source transitions affect value learning only through the TBM-weighted target-aligned critic objective, while value fitting and policy extraction follow target-domain data.
The algorithm of our method is presented in Appendix~\ref{app:algorithm details}.

\subsection{Theoretical Analysis} 
\label{sec:theoretical_analysis}

Incorporating source-domain data may introduce bias in target-domain value estimation due to discrepancies in dynamics, which can adversely affect policy learning. Here, we provide a theoretical justification showing that the proposed TBM-based reweighting scheme reduces and controls this bias by down-weighting transitions that are inconsistent with target-domain dynamics.

We first define the \emph{target Bellman error} for a source domain transition $\xi=(\textbf{s},\textbf{a},r,\textbf{s}')$ as:
\begin{equation}
\delta_{\mathrm{tar}}(\xi)
=
\left|
\left(r + \gamma V(\mathbf{z}'_{s})\right) - r_{\mathrm{tar}} + \gamma V(\mathbf{z}'_{s,\mathrm{tar}})
\right|.
\end{equation}
Here, $r_{\mathrm{tar}}$ and $\mathbf{z}'_{s,\mathrm{tar}}$ denote the reward and next-state latent representation that would be obtained if the same state-action pair $(\textbf{s},\textbf{a})$ were evaluated under target-domain dynamics. Therefore, $\delta_{\mathrm{tar}}(\xi)$ measures the discrepancy between the Bellman backup induced by a source transition and the one that would be obtained under the target domain, characterizing the bias introduced when using source data for target value estimation.

Since $r_{\mathrm{tar}}$ and $\mathbf{z}'_{s,\mathrm{tar}}$ are not directly observable for source-domain data, we approximate them using the learned target-domain predictor $f_{\mathrm{ref}}$, leading to the following target Bellman mismatch (TBM): $\mathrm{TBM}(\xi)
=
\left|
(r+\gamma V(\mathbf{z}'_{s}))
-
(\hat r+\gamma V(\hat{\mathbf{z}}'_{s}))
\right|$.

\paragraph{Theorem 1 (Bias Control via TBM Weighting).}
Assume that:
(i) the target-domain predictor $f_{\mathrm{ref}}$ is $\epsilon$-accurate, i.e.,
\[
|\hat r - r_{\mathrm{tar}}| \le \epsilon, 
\quad
\|\hat{\mathbf{z}}'_{s} - \mathbf{z}'_{s,\mathrm{tar}}\| \le \epsilon,
\]
and
(ii) the value function $V$ is $L$-Lipschitz.
Then the TBM provides an upper bound of the target Bellman error (We provide detailed proof in Appendix~\ref{app:proof}):
\begin{equation}
\delta_{\mathrm{tar}}(\xi)
\le
\mathrm{TBM}(\xi)
+
(1 + \gamma L)\epsilon.
\end{equation}
Furthermore, the weight of source data is computed by Eq.~\eqref{eq:weight}. Compared to uniform weighting, the proposed scheme assigns smaller weights to transitions with large TBM. Since TBM serves as a proxy for the target Bellman error $\delta_{\mathrm{tar}}(\xi)$, this effectively down-weights transitions that are highly inconsistent with target-domain dynamics, thereby reducing the bias in target-domain value estimation and improving policy learning.

\section{Experiments}

We evaluate TABB across a diverse set of cross-domain offline RL tasks, and we focus on answering the following questions: 
(\romannumeral 1) Can TABB outperform existing cross-domain offline RL methods across locomotion and manipulation tasks under diverse dynamics shifts? (\textbf{RQ1})
(\romannumeral 2) How does each key component of TABB affect the final performance? (\textbf{RQ2})
(\romannumeral 3) Is TABB robust to different dynamics-shift intensities and heterogeneous source-target dataset qualities? (\textbf{RQ3})
(\romannumeral 4) Does the proposed TBM provide a more reliable measure of source transition transferability than transition-similarity-based criteria? (\textbf{RQ4})

\subsection{Experiment Setup}
\textbf{Tasks and Datasets.}
Following previous works~\cite{lyu2024odrl,guo2025mobody}, our benchmark tasks cover two categories: locomotion~\cite{todorov2012mujoco} and manipulation~\cite{rajeswaran2017learning}. For locomotion, we evaluate on HalfCheetah, Hopper, Walker2d, and Ant with there types of dynamics shifts: morphology, kinematics, and friction shifts. We use the standard D4RL~\cite{fu2020d4rl} datasets as source-domain data, while collecting target-domain datasets from the corresponding dynamics-shifted environments using the same data collection protocol. For manipulation, we evaluate on Pen and Door, and introduce kinematics and morphology shifts in the target domains. Across all tasks, each source dataset contains 1 million transitions~\cite{haarnoja2018soft}, whereas each target dataset contains only 5K transitions. More experimental details can be found in Appendix~\ref{app:exp}.

\textbf{Baselines.}
We compare TABB against a range of cross-domain offline RL methods, including DARA~\cite{liu2022dara}, BOSA~\cite{liu2024beyond}, SRPO~\cite{kumar2019stabilizing}, OTDF~\cite{lyu2025cross}, DROCO~\cite{qiao2025dual} and MOBODY~\cite{guo2025mobody}. We also evaluate IQL~\cite{kostrikov2021offline} trained directly on the mixed offline dataset containing both source- and target-domain transitions.

\subsection{Main Results}
\label{sec:main_results}
\textbf{Results on locomotion tasks.} Table~\ref{tab:mujoco} presents the results of TABB and the baselines on MuJoCo environments under morphology, kinematics, and friction shifts, each result is reported with  normalized returns over 5 random seeds. TABB achieves the best performance on 9 out of 12 tasks. Overall, it outperforms MOBODY, the strongest baseline, by \textbf{29}\%, demonstrating its effectiveness across multiple environments and diverse dynamics shifts.

\begin{table}[t]
\scriptsize
    \centering
    \caption{Results on MuJoCo tasks with different dynamics shifts.}
    \label{tab:mujoco}
    \setlength{\tabcolsep}{3.0pt}
    \renewcommand{\arraystretch}{1.05}
    \resizebox{\linewidth}{!}{%
    \begin{tabular}{c|c|ccccccc|@{\hspace{4pt}}c@{\hspace{4pt}}}
    \toprule
        Env & Shift & IQL & DARA & BOSA & SRPO & OTDF & DROCO & MOBODY & TABB \\
        \midrule
        \multirow{3}{*}{Halfcheetah} & Morphology & 30.0$\pm$1.6 & 26.6$\pm$3.3 & 19.3$\pm$3.5 & 41.3$\pm$0.4 & 39.1$\pm$2.3 & 41.0$\pm$0.7 & \tbsecond{42.1$\pm$0.4} & \tbbest{44.0$\pm$0.3} \\
         & Kinematic & 12.3$\pm$1.2 & 10.6$\pm$1.2 & 8.3$\pm$1.2 & 16.8$\pm$4.2 & \tbsecond{40.2$\pm$0.0} & 13.5$\pm$3.8 & 35.1$\pm$6.3 & \tbbest{42.0$\pm$0.3} \\
         & Friction & 45.0$\pm$6.8 & 40.1$\pm$7.9 & 44.1$\pm$9.1 & 5.1$\pm$2.0 & 38.9$\pm$2.0 & 56.9$\pm$2.2 & \tbsecond{59.2$\pm$4.9} & \tbbest{63.5$\pm$7.0} \\
        \midrule
        \multirow{3}{*}{Hopper} & Morphology & \tbsecond{13.5$\pm$0.2} & 13.5$\pm$0.4 & 13.2$\pm$0.3 & 13.4$\pm$0.1 & 11.0$\pm$0.9 & 13.0$\pm$0.4 & 13.1$\pm$0.6 & \tbbest{63.5$\pm$1.0} \\
         & Kinematic & 58.7$\pm$8.4 & 43.9$\pm$15.2 & 12.3$\pm$6.6 & 65.4$\pm$1.5 & \tbsecond{65.6$\pm$1.9} & 45.4$\pm$22.3 & 62.4$\pm$7.2 & \tbbest{66.2$\pm$0.7} \\
         & Friction & 7.9$\pm$0.0 & 7.9$\pm$0.1 & 7.9$\pm$0.1 & 7.8$\pm$0.3 & 8.0$\pm$0.1 & \tbsecond{8.2$\pm$0.1} & 8.1$\pm$0.0 & \tbbest{8.2$\pm$0.0} \\
        \midrule
        \multirow{3}{*}{Walker2d} & Morphology & 23.0$\pm$4.7 & 23.3$\pm$3.3 & 6.2$\pm$2.9 & 24.7$\pm$1.7 & \tbbest{50.5$\pm$5.8} & 32.8$\pm$6.6 & 24.7$\pm$6.4 & \tbsecond{47.2$\pm$3.7} \\
         & Kinematic & 34.3$\pm$9.8 & 35.2$\pm$22.5 & 14.3$\pm$11.2 & 39.0$\pm$6.7 & 49.6$\pm$18.0 & 46.9$\pm$8.9 & \tbbest{56.4$\pm$4.1} & \tbsecond{50.2$\pm$5.4} \\
         & Friction & 5.4$\pm$0.0 & 5.4$\pm$0.3 & 10.1$\pm$4.9 & 4.9$\pm$0.7 & 13.8$\pm$6.5 & 27.0$\pm$6.9 & \tbsecond{27.4$\pm$3.9} & \tbbest{40.8$\pm$3.5} \\
        \midrule
        \multirow{3}{*}{Ant} & Morphology & 38.7$\pm$3.8 & \tbsecond{41.3$\pm$1.8} & 18.2$\pm$1.9 & 40.6$\pm$2.1 & 39.4$\pm$1.7 & 40.2$\pm$2.3 & 40.9$\pm$0.9 & \tbbest{41.8$\pm$0.7} \\
         & Kinematic & 50.0$\pm$5.6 & 42.3$\pm$7.6 & 20.9$\pm$2.6 & 50.5$\pm$6.7 & \tbbest{55.4$\pm$0.0} & 50.6$\pm$8.3 & 50.6$\pm$4.6 & \tbsecond{53.9$\pm$2.5} \\
         & Friction & 7.8$\pm$0.3 & 7.8$\pm$0.1 & 9.1$\pm$0.9 & 11.7$\pm$1.9 & 13.4$\pm$3.7 & 23.7$\pm$8.5 & \tbsecond{31.2$\pm$5.6} & \tbbest{62.9$\pm$2.5} \\
        \midrule
        Total &  & 326.6 & 297.9 & 183.9 & 321.2 & 424.9 & 399.2 & \tbsecond{451.2} & \tbbest{584.2} \\
        \bottomrule
    \end{tabular}%
    }
\end{table}

\begin{table}[t]
\scriptsize
    \centering
    \caption{Results on Adroit tasks with different dynamics shifts.}
    \label{tab:adroit}
    \setlength{\tabcolsep}{3.0pt}
    \renewcommand{\arraystretch}{1.05}
    \resizebox{\linewidth}{!}{%
    \begin{tabular}{c|c|c|ccccccc|@{\hspace{4pt}}c@{\hspace{4pt}}}
    \toprule
        Env & Type & Level & IQL & DARA & BOSA & SRPO & OTDF & DROCO & MOBODY & TABB \\
        \midrule
        \multirow{4}{*}{Pen} & \multirow{2}{*}{kin-broken-jnt} & M & 24.3$\pm$15.5 & 38.6$\pm$3.4 & 30.6$\pm$9.0 & 23.4$\pm$2.4 & 41.2$\pm$2.9 & \tbsecond{50.4$\pm$9.2} & 37.7$\pm$4.5 & \tbbest{83.9$\pm$20.0} \\
         &  & H & 7.7$\pm$3.5 & 9.4$\pm$6.1 & 7.2$\pm$2.0 & 8.1$\pm$5.8 & 12.6$\pm$10.4 & \tbsecond{18.9$\pm$15.8} & 13.7$\pm$6.3 & \tbbest{40.4$\pm$2.6} \\
         & \multirow{2}{*}{morph-shrink-finger} & M & 13.8$\pm$4.9 & 8.7$\pm$3.1 & 10.7$\pm$6.7 & 7.6$\pm$3.7 & 14.0$\pm$6.5 & 10.8$\pm$1.1 & \tbsecond{16.5$\pm$10.5} & \tbbest{30.1$\pm$6.0} \\
         &  & H & 32.2$\pm$1.1 & 22.2$\pm$3.9 & 11.8$\pm$6.6 & 8.8$\pm$6.8 & 27.6$\pm$14.8 & 18.9$\pm$3.9 & \tbsecond{37.8$\pm$1.2} & \tbbest{38.8$\pm$5.9} \\
        \midrule
        \multirow{4}{*}{Door} & \multirow{2}{*}{kin-broken-jnt} & M & 37.4$\pm$12.8 & 20.2$\pm$5.3 & 25.4$\pm$22.0 & 0.7$\pm$1.3 & 32.9$\pm$17.5 & \tbsecond{45.3$\pm$13.9} & 39.3$\pm$3.7 & \tbbest{53.1$\pm$6.7} \\
         &  & H & 56.0$\pm$7.7 & 58.2$\pm$9.9 & 30.6$\pm$26.9 & 0.6$\pm$1.3 & 48.1$\pm$18.4 & 46.9$\pm$6.9 & \tbsecond{61.6$\pm$9.8} & \tbbest{73.8$\pm$8.6} \\
         & \multirow{2}{*}{morph-shrink-finger} & M & 60.7$\pm$12.8 & 50.3$\pm$4.8 & 41.6$\pm$6.0 & 1.4$\pm$1.0 & 56.8$\pm$2.4 & 61.6$\pm$11.2 & \tbbest{63.7$\pm$9.5} & \tbsecond{63.4$\pm$5.0} \\
         &  & H & 68.6$\pm$8.3 & 44.2$\pm$7.2 & 27.0$\pm$8.6 & 0.8$\pm$0.6 & 64.3$\pm$7.9 & \tbsecond{69.1$\pm$3.0} & 62.9$\pm$5.3 & \tbbest{69.9$\pm$4.4} \\
        \midrule
        Total &  &  & 300.7 & 251.8 & 184.9 & 51.4 & 297.5 & 321.9 & \tbsecond{333.2} & \tbbest{453.4} \\
        \bottomrule
    \end{tabular}%
    }
\end{table}

Notably, TABB achieves significant improvements over IQL on challenging tasks such as Hopper-Morphology, Ant-Friction, and Walker2d-Friction, where strong dynamics shifts substantially alter the consequences of actions and cause most baselines to remain close to the IQL performance level.
indicating that similarity-based source reuse may introduce source transitions that are visually or structurally similar but fail to provide effective supervision for policy learning. 
In contrast, TABB evaluates source transitions by their effect on target-domain Bellman updates, which helps suppress target-inconsistent source supervision and leads to substantial performance gains.
In the Walker2d environment, TABB slightly underperforms OTDF and MOBODY, possibly because the morphology and kinematic tasks are more limited by the coverage of different target-domain locomotion phases. Such phases can be better preserved or augmented through distribution alignment or target-dynamics generation, while TABB mainly evaluates the Bellman contribution of individual source transitions.

\textbf{Results on manipulation tasks.}
Table~\ref{tab:adroit} reports the results on Adroit manipulation tasks under kinematic and morphology shifts. 
TABB achieves the best performance on 7 out of 8 tasks and ranks second on the remaining one, improving the total score from 333.2 of the strongest baseline (MOBODY) to 453.4, corresponding to a 36.1\% improvement.

\subsection{Ablation Study}
\enlargethispage{\baselineskip}

\begin{wraptable}{r}{0.45\linewidth}
\vspace{-8pt}
\centering
\scriptsize
\caption{The ablation study.}
\label{tab:tabb_ablation}
\setlength{\tabcolsep}{2.0pt}
\resizebox{\linewidth}{!}{
\begin{tabular}{@{}llccc@{}}
\toprule
Env. & Shift & -TBM & SrcActor & TABB \\
\midrule
Ant & morph & $38.7{\pm}3.8$ & \tbbest{$41.7{\pm}0.1$}  & \tbsecond{$40.9{\pm}1.0$} \\
Halfcheetah & kin & \tbsecond{$12.3{\pm}1.2$}  & $9.3{\pm}10.5$ & \tbbest{$41.7{\pm}0.1$} \\
Walker2d & morph & $23.0{\pm}4.7$ & \tbsecond{$28.3{\pm}13.8$} 
 & \tbbest{$43.4{\pm}5.3$}\\
Hopper & kin &  \tbsecond{$58.7{\pm}8.4$}  & $23.4{\pm}5.7$ & \tbbest{$64.6{\pm}1.6$}\\
\bottomrule
\end{tabular}
}
\end{wraptable}

We further conduct an ablation study to investigate the contribution of key design of TABB. Specifically, we have 2 variants: 
(1)\textbf{-TBM:} removing TBM from TABB, which is equivalent to using IQL to train the policy directly on the source and target domain data. 
(2) \textbf{SrcActor}: building upon the critic update with TBM-weighted source transitions, we further leverage source data to train the actor.

The ablation results are reported in Table \ref{tab:tabb_ablation}. From these results, we can find that:
(1) TBM provides the basis for effective target-domain policy learning.
As shown in Table~\ref{tab:tabb_ablation}, removing TBM leads to clear performance degradation on most tasks, especially on HalfCheetah-Kinematic and Walker2d-Morphology. 
Without TBM, source transitions are directly mixed with target data and participate in target-domain Bellman updates without transferability evaluation. 
This may bias target-domain value learning and further harm the learned target policy, since similar state-action pairs can induce different transition outcomes across domains. 
The large gap between \textbf{-TBM} and TABB indicates that evaluating and weighting source transitions according to TBM is beneficial.
(2) Directly using source transitions for policy updates degrades policy performance.
As shown in Table~\ref{tab:tabb_ablation}, SrcActor leads to severe performance degradation on most tasks. 
This is because forcing the actor to imitate source actions may introduce behavior patterns that are not supported by the target domain, thereby increasing the risk of negative transfer. 
The degradation of \textbf{SrcActor} suggests that source data are beneficial when used as target-aligned critic supervision, but can be harmful when used for policy improvement.

\subsection{Robustness Analysis}

\begin{wrapfigure}[13]{r}{0.5\linewidth}
    \vspace{-40pt}
    \centering
    \includegraphics[width=\linewidth]{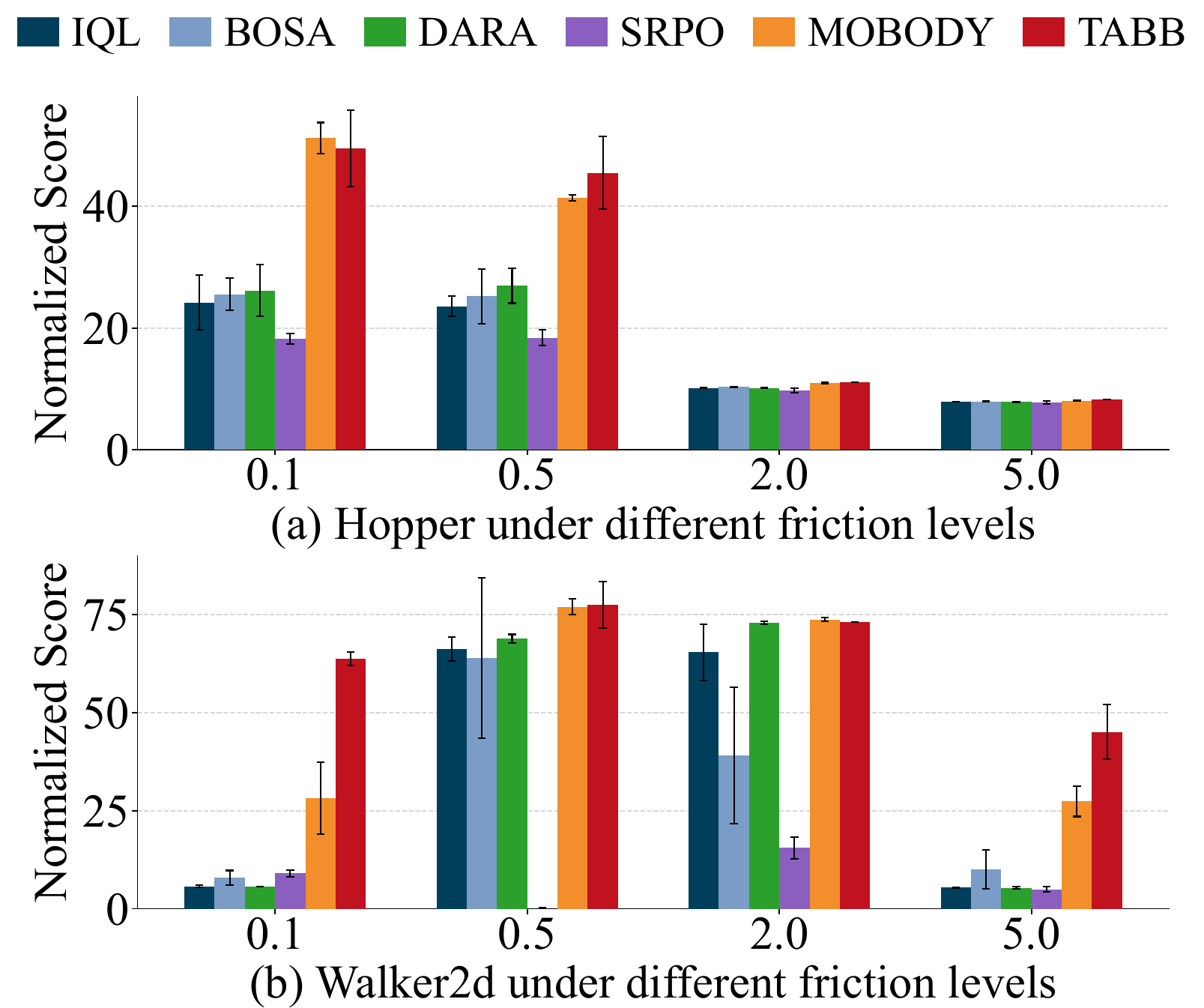}
    \vspace{-10pt}
    \caption{Results under different levels.}
    \label{fig:robust_level}
\end{wrapfigure}

We evaluate the robustness of TABB from two perspectives: varying dynamics-shift intensities and heterogeneous source-target data quality.

\begin{table*}[t]
\scriptsize
    \centering
    \caption{Results on Halfcheetah and Ant under varying source and target qualities.}
    \label{tab:robust_qualities}
    \setlength{\tabcolsep}{3.0pt}
    \renewcommand{\arraystretch}{1.05}
    \resizebox{\textwidth}{!}{%
    \begin{tabular}{c|c|c|c|cccccc}
    \toprule
        Env & Shift & Src & Tar & IQL & DARA & BOSA & DROCO & MOBODY & TABB \\
        \midrule
        \multirow{9}{*}{Halfcheetah} & \multirow{9}{*}{Morph} & \multirow{3}{*}{medium} & medium & 30.00$\pm$1.60 & 26.60$\pm$3.30 & 19.30$\pm$3.50 & 41.06$\pm$0.33 & \tbsecond{42.16$\pm$0.35} & \tbbest{43.99$\pm$0.24} \\
         &  &  & medium-expert & 31.80$\pm$1.10 & 32.00$\pm$0.70 & \tbsecond{33.60$\pm$1.10} & 32.85$\pm$0.38 & 29.94$\pm$1.24 & \tbbest{43.13$\pm$0.24} \\
         &  &  & expert & 8.50$\pm$1.00 & \tbsecond{9.30$\pm$1.60} & 7.90$\pm$0.80 & 8.16$\pm$1.39 & 8.41$\pm$1.37 & \tbbest{59.60$\pm$6.70} \\
         &  & \multirow{3}{*}{medium-replay} & medium & 30.80$\pm$4.40 & 35.60$\pm$0.70 & 35.00$\pm$4.60 & 20.49$\pm$1.86 & \tbsecond{39.21$\pm$0.96} & \tbbest{43.95$\pm$0.42} \\
         &  &  & medium-expert & 12.90$\pm$2.20 & 16.90$\pm$4.10 & 19.90$\pm$5.50 & 8.56$\pm$1.60 & \tbsecond{21.38$\pm$4.93} & \tbbest{43.01$\pm$0.21} \\
         &  &  & expert & \tbsecond{5.90$\pm$1.70} & 3.70$\pm$2.70 & 2.40$\pm$1.90 & 4.32$\pm$0.91 & 4.70$\pm$2.01 & \tbbest{53.99$\pm$7.43} \\
         &  & \multirow{3}{*}{medium-expert} & medium & 41.50$\pm$0.10 & 40.30$\pm$1.20 & 41.30$\pm$0.30 & 40.92$\pm$0.73 & \tbsecond{41.99$\pm$0.34} & \tbbest{44.03$\pm$0.16} \\
         &  &  & medium-expert & 25.80$\pm$2.00 & 30.60$\pm$2.80 & \tbsecond{32.10$\pm$0.80} & 30.22$\pm$1.58 & 29.59$\pm$1.94 & \tbbest{42.55$\pm$0.64} \\
         &  &  & expert & 7.80$\pm$1.30 & 8.30$\pm$1.30 & 9.10$\pm$0.80 & \tbsecond{9.29$\pm$0.53} & 7.36$\pm$1.30 & \tbbest{46.11$\pm$8.05} \\
        \midrule
        \multirow{9}{*}{Ant} & \multirow{9}{*}{Kinematic} & \multirow{3}{*}{medium} & medium & 50.00$\pm$5.60 & 42.30$\pm$7.60 & 20.90$\pm$2.60 & 46.04$\pm$3.72 & \tbsecond{50.56$\pm$4.64} & \tbbest{53.87$\pm$2.51} \\
         &  &  & medium-expert & 57.80$\pm$7.20 & 54.10$\pm$3.80 & 31.70$\pm$7.00 & 50.86$\pm$8.78 & \tbsecond{59.12$\pm$6.84} & \tbbest{65.10$\pm$11.14} \\
         &  &  & expert & 59.60$\pm$18.50 & 54.20$\pm$11.30 & 45.40$\pm$8.60 & 46.89$\pm$16.98 & \tbsecond{76.04$\pm$13.16} & \tbbest{83.67$\pm$13.33} \\
         &  & \multirow{3}{*}{medium-replay} & medium & 43.70$\pm$4.60 & 42.00$\pm$5.40 & 19.00$\pm$1.80 & 41.85$\pm$3.93 & \tbsecond{44.98$\pm$6.48} & \tbbest{54.44$\pm$5.41} \\
         &  &  & medium-expert & 36.50$\pm$5.90 & 36.00$\pm$6.70 & 19.10$\pm$1.60 & 29.26$\pm$5.16 & \tbsecond{37.00$\pm$5.98} & \tbbest{63.98$\pm$5.10} \\
         &  &  & expert & 24.40$\pm$4.80 & 22.10$\pm$0.40 & 19.50$\pm$0.80 & 22.88$\pm$0.52 & \tbsecond{24.62$\pm$3.29} & \tbbest{87.05$\pm$9.40} \\
         &  & \multirow{3}{*}{medium-expert} & medium & \tbsecond{49.50$\pm$4.10} & 44.70$\pm$4.30 & 19.00$\pm$8.00 & 44.57$\pm$8.68 & 47.72$\pm$4.86 & \tbbest{51.70$\pm$3.08} \\
         &  &  & medium-expert & 37.20$\pm$2.00 & 33.30$\pm$7.00 & 6.40$\pm$2.50 & 29.71$\pm$8.17 & \tbsecond{50.93$\pm$5.47} & \tbbest{67.91$\pm$3.82} \\
         &  &  & expert & 18.70$\pm$8.10 & 17.80$\pm$23.60 & 14.50$\pm$9.00 & 25.55$\pm$21.58 & \tbsecond{36.86$\pm$21.61} & \tbbest{90.34$\pm$7.50} \\
        \bottomrule
    \end{tabular}%
    }
\end{table*}

\textbf{Varying dynamics-shift intensities.}
We construct cross-domain tasks with varying friction levels in the Hopper and Walker2D environments. 
The friction level is selected from $\{0.1, 0.5, 2.0, 5.0\}$, covering a broad range of dynamics variations. The results in Figure~\ref{fig:robust_level} show that TABB achieves the best or comparable performance across all friction levels, demonstrating robust performance under varying contact-dynamics shifts.
In Walker2d, TABB shows strong advantages at friction levels 0.1 and 5.0, substantially outperforming the strongest baseline. 
In Hopper, the advantage is less pronounced, possibly because its simpler single-leg structure makes friction-induced transition changes more localized, allowing transition-level similarity to identify useful source data to some extent. Nevertheless, TABB remains competitive, demonstrating the robustness of TBM-based source reuse under varying friction shifts.

\textbf{Heterogeneous source-target data qualities.}
We conduct experiments on HalfCheetah-Morphology and Ant-Kinematic under varying source and target data qualities. The source datasets include medium, medium-replay, and medium-expert qualities, while the target datasets include medium, medium-expert, and expert qualities.
As shown in Table~\ref{tab:robust_qualities},
TABB achieves the best performance across all 18 source-target data-quality combinations, demonstrating strong robustness to heterogeneous dataset qualities. The advantage is especially pronounced when the target data are expert-level. On HalfCheetah-Morphology, TABB obtains 59.60, 53.99, and 46.11 with medium, medium-replay, and medium-expert source data, respectively, while most baselines remain below 10. On Ant-Kinematic, TABB reaches 87.05 and 90.34 with medium-replay and medium-expert source data, substantially outperforming the strongest baselines. 
These results indicate that when source and target datasets differ substantially in behavior quality, transition-similarity can become unreliable, whereas TABB identifies source transitions through their contribution to target-domain Bellman backup, leading to more robust source reuse.

\subsection{Further Investigations}

\begin{wrapfigure}{r}{0.40\linewidth}
    \centering
    \vspace{-30pt}
    \includegraphics[width=\linewidth]{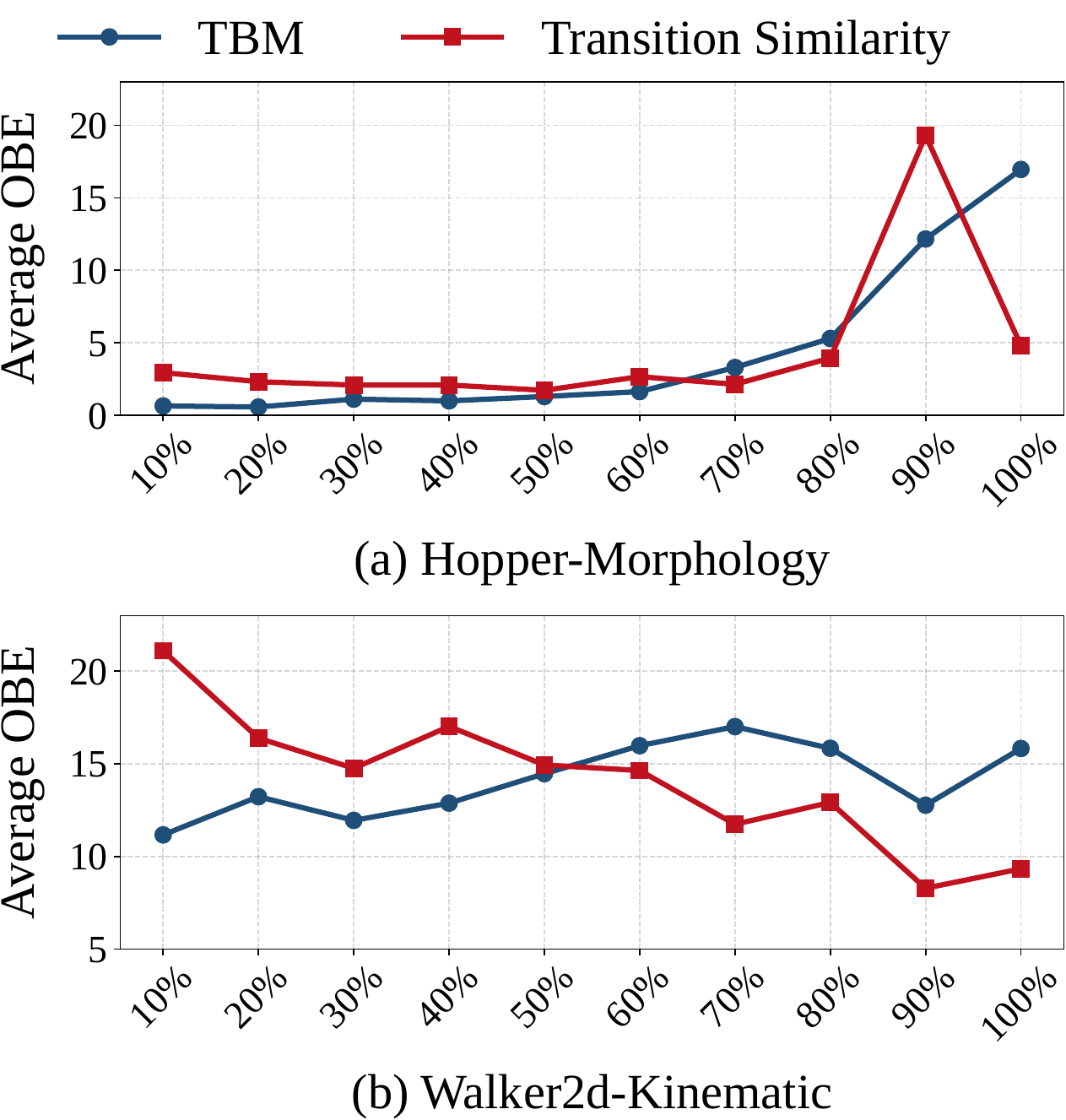}
    \vspace{-15pt}
    \caption{Oracle Bellman Error of source transitions ranked by TBM and transition similarity. }
    \vspace{-5pt}
    \label{fig:tbm_otdf}
    \vspace{-6pt}
\end{wrapfigure}

TABB uses TBM to estimate the target-domain Bellman target for each source transition, and measures its mismatch with the original source Bellman target. The resulting mismatch is then used to reweight source transitions according to their consistency with target-domain Bellman learning. To empirically examine whether the estimated TBM accurately reflects the true source-to-target Bellman error, we conduct a diagnostic experiment.
Specifically, we sample 10K source transitions and replay their state-action inputs once in the target environment. This online evaluation is not used for training; it only provides the target-domain reward and next state for computing an oracle Bellman mismatch
oracle Bellman error.
We then compute both TBM and a transition-similarity score (we use OT distance in~\cite{lyu2025cross}) to evaluate how well each criteria aligns with the corresponding oracle.
We rank 10K source transitions from low to high according to their scores and divide the ranked transitions into ten equal-sized percentile groups (i.e., top 0–10\%, 10–20\%, …), reporting the corresponding average Bellman error for each group. Intuitively, an effective metric should assign lower scores to source transitions that are more consistent with the target domain; accordingly, higher-ranked groups are expected to exhibit smaller average Bellman errors. Therefore, the curve is expected to show a generally increasing trend as the score increases.
The results are illustrated in Figure~\ref{fig:tbm_otdf}, we can observe that on both Hopper-Morphology and Walker2d-Kinematic, the Bellman error increase with the TBM score. In contrast, on Hopper-Morphology, the Bellman error initially increases as the OT score grows, but unexpectedly decreases in the high-score region. In Walker2d-Kinematic, the Bellman error even decreases as the OT score increases. 
Since a larger oracle Bellman error indicates that the source transition provides a less reliable Bellman target for the target domain, the results in Figure~\ref{fig:tbm_otdf} show that TBM better identifies source transitions that may be harmful to target-domain value learning. This explains why TBM-based weighting is more effective than transition-similarity-based weighting and further supports the performance gains of TABB observed in the Section~\ref{sec:main_results}.

\section{Conclusion}

In this paper, we propose a novel method for cross-domain offline reinforcement learning that filters and reweights source data based on the Bellman target mismatch. Specifically, we first encode both source and target data into a latent space, where we estimate the discrepancy between their Bellman targets. Based on this discrepancy, we assign weights to source transitions and incorporate the weighted source data into the training of the target policy.
Extensive experiments demonstrate that our method, TABB, can effectively improve the performance of cross-domain offline RL.

\textbf{Limitations and Future Directions.} 
Although TABB achieves strong performance across diverse environments and dynamics shifts, our current study focuses on domains with shared state and action spaces. Extending TABB to more heterogeneous cross-domain settings, such as different observation spaces or embodiment structures, remains an important future direction.

\bibliographystyle{unsrtnat}
\bibliography{references}

\newpage
\appendix

\section{Proof}
\label{app:proof}
In this section, we formally present the proof of Theorem 1.

We decompose the target Bellman error as:
\begin{align}
\delta_{\mathrm{tar}}(\xi)
&=
\left|
(r_{\mathrm{tar}} - \hat r)
+
\gamma \left(
V(\mathbf{z}'_{s,\mathrm{tar}})
-
V(\hat{\mathbf{z}}'_{s})
\right)
\right. \nonumber \\
&\quad \left.
+
\left(
\hat r + \gamma V(\hat{\mathbf{z}}'_{s})
-
(r + \gamma V(\mathbf{z}'_{s}))
\right)
\right|.
\end{align}
Applying the triangle inequality yields:
\begin{equation}
\delta_{\mathrm{tar}}(\xi)
\le
|\hat r - r_{\mathrm{tar}}|
+
\gamma |V(\mathbf{z}'_{s,\mathrm{tar}}) - V(\hat{\mathbf{z}}'_{s})|
+
\mathrm{TBM}(\xi).
\end{equation}
Using the Lipschitz continuity of $V$, we further obtain:
\begin{equation}
|V(\mathbf{z}'_{s,\mathrm{tar}}) - V(\hat{\mathbf{z}}'_{s})|
\le
L \|\mathbf{z}'_{s,\mathrm{tar}} - \hat{\mathbf{z}}'_{s}\|.
\end{equation}
Combining the above inequalities with the $\epsilon$-accuracy assumption completes the proof.

\section{Experiment Details}
\label{app:exp}
In this section, we provide a detailed description of the experimental environment design, including the source-domain and target-domain datasets, as well as the implementation details of morphology, kinematic, friction, and Adroit dynamics shifts.

\subsection{Datasets}
\textbf{Source-domain datasets.}
For the locomotion experiments, we directly adopt the MuJoCo datasets from D4RL, which are widely used benchmarks in offline RL. We consider four locomotion environments: HalfCheetah, Hopper, Walker2d, and Ant.

For the dexterous manipulation experiments, we use the Adroit domain from D4RL as the source domain. We evaluate two manipulation tasks, Pen and Door, both of which require controlling a high-dimensional Shadow Hand.

\textbf{Target-domain datasets.}
To evaluate the effectiveness of our method under dynamics shifts, we design two groups of target-domain benchmarks.
The first group is based on the MuJoCo locomotion environments used in the source domain, including HalfCheetah, Hopper, Walker2d, and Ant. We consider three types of dynamics shifts: morphology, kinematic, and friction shifts. The morphology shift changes the agent's body structure between the source and target domains, the kinematic shift simulates partial joint failure by restricting selected joint rotation ranges, and the friction shift changes contact friction coefficients while keeping the robot appearance unchanged.

The second group is based on Adroit dexterous manipulation tasks. We evaluate Pen and Door under two types of dynamics shifts: kinematic broken-joint and morphology shrink-finger. The kinematic broken-joint shift constrains the rotation ranges of selected hand joints, while the morphology shrink-finger shift shortens selected finger phalanges. For each shift type, we consider two difficulty levels: medium and hard.

In the modified MuJoCo locomotion environments, we follow the data collection protocol established by D4RL and ODRL. For each target task, we use 5000 transitions to evaluate policy learning under limited target-domain data. For Adroit, the target-domain datasets also follow the low-data off-dynamics setting with 5000 transitions per task.

\textbf{Evaluation metric.}
We employ the standard D4RL normalized score to measure the agent's performance in the target domain:
\begin{equation}
    \text{NS}=\frac{J_{\pi}-J_{\text{r}}}{J_{\text{e}}-J_{\text{r}}}\times 100\%,
\end{equation}
where $J_{\pi}$ denotes the return obtained by the learned policy in the target domain, while $J_{\text{r}}$ and $J_{\text{e}}$ denote the returns of a random policy and an expert policy in the target domain, respectively.

\subsection{Friction Shift}
In the target domain, we realize friction shifts by modifying the \texttt{friction} attribute in the MuJoCo \texttt{geom} elements. The friction coefficients in the target domain are scaled by $0.1$, $0.5$, $2.0$, or $5.0$ relative to the source-domain coefficients. This shift changes the contact dynamics but does not change the visual morphology of the agent.

The following Hopper-friction-5.0 snippet illustrates the modification:
\begin{lstlisting}[language=Python]
# Hopper-friction-5.0
<geom friction="4.5" fromto="0 0 1.45 0 0 1.05" name="torso_geom" size="0.05" type="capsule"/>
<geom friction="4.5" fromto="0 0 1.05 0 0 0.6" name="thigh_geom" size="0.05" type="capsule"/>
<geom friction="4.5" fromto="0 0 0.6 0 0 0.1" name="leg_geom" size="0.04" type="capsule"/>
<geom friction="10.0" fromto="-0.13 0 0.1 0.26 0 0.1" name="foot_geom" size="0.06" type="capsule"/>
\end{lstlisting}

\subsection{Morphology Shift}
In the target domain, we modify the morphological structure of the agents. Specifically:

\textbf{\textit{HalfCheetah-morph}}: we adjust the sizes of the back thigh and the forward thigh.
\begin{lstlisting}[language=Python]
# back thigh
<geom fromto="0 0 0 -0.0001 0 -0.0001" name="bthigh" size="0.046" type="capsule"/>
<body name="bshin" pos="-0.0001 0 -0.0001">
# front thigh
<geom fromto="0 0 0 0.0001 0 0.0001" name="fthigh" size="0.046" type="capsule"/>
<body name="fshin" pos="0.0001 0 0.0001">
\end{lstlisting}

\textbf{\textit{Hopper-morph}}: we increase the size of the agent's head.
\begin{lstlisting}[language=Python]
# head size
<geom friction="0.9" fromto="0 0 1.45 0 0 1.05" name="torso_geom" size="0.125" type="capsule"/>
\end{lstlisting}

\textbf{\textit{Walker2d-morph}}: we modify the thigh on the right leg.
\begin{lstlisting}[language=Python]
# right leg
<body name="thigh" pos="0 0 1.05">
<joint axis="0 -1 0" name="thigh_joint" pos="0 0 1.05" range="-150 0" type="hinge"/>
<geom friction="0.9" fromto="0 0 1.05 0 0 1.045" name="thigh_geom" size="0.05" type="capsule"/>
<body name="leg" pos="0 0 0.35">
  <joint axis="0 -1 0" name="leg_joint" pos="0 0 1.045" range="-150 0" type="hinge"/>
  <geom friction="0.9" fromto="0 0 1.045 0 0 0.3" name="leg_geom" size="0.04" type="capsule"/>
  <body name="foot" pos="0.2 0 0">
    <joint axis="0 -1 0" name="foot_joint" pos="0 0 0.3" range="-45 45" type="hinge"/>
    <geom friction="0.9" fromto="-0.0 0 0.3 0.2 0 0.3" name="foot_geom" size="0.06" type="capsule"/>
  </body>
</body>
</body>
\end{lstlisting}

\textbf{\textit{Ant-morph}}: we reduce the sizes of the agent's two front legs.
\begin{lstlisting}[language=Python]
# front leg 1
<geom fromto="0.0 0.0 0.0 0.1 0.1 0.0" name="left_ankle_geom" size="0.08" type="capsule"/>
# front leg 2
<geom fromto="0.0 0.0 0.0 -0.1 0.1 0.0" name="right_ankle_geom" size="0.08" type="capsule"/>
\end{lstlisting}

\subsection{Kinematic Shift}
We realize cross-domain kinematic shift by adjusting the rotational ranges of specific agent joints. Specifically:

\textbf{\textit{HalfCheetah-kinematic}}: The rotation angle of the joint on the thigh of the robot's back leg is adjusted from $[-0.52, 1.05]$ to $[-0.0052, 0.0105]$.
\begin{lstlisting}[language=Python]
# broken back thigh joint
<joint axis="0 1 0" damping="6" name="bthigh" pos="0 0 0" range="-.0052 .0105" stiffness="240" type="hinge"/>
\end{lstlisting}

\textbf{\textit{Hopper-kinematic}}: The rotation angle of the head joint is adjusted from $[-150, 0]$ to $[-0.15, 0]$, and the rotation angle of the foot joint is adjusted from $[-45, 45]$ to $[-18, 18]$.
\begin{lstlisting}[language=Python]
# broken head joint
<joint axis="0 -1 0" name="thigh_joint" pos="0 0 1.05" range="-0.15 0" type="hinge"/>
# broken foot joint
<joint axis="0 -1 0" name="foot_joint" pos="0 0 0.1" range="-18 18" type="hinge"/>
\end{lstlisting}

\textbf{\textit{Walker2d-kinematic}}: The rotational range of the right foot joint is adjusted from $[-45, 45]$ to $[-0.45, 0.45]$.
\begin{lstlisting}[language=Python]
# broken right foot joint
<joint axis="0 -1 0" name="foot_joint" pos="0 0 0.1" range="-0.45 0.45" type="hinge"/>
\end{lstlisting}

\textbf{\textit{Ant-kinematic}}: The rotation angles of the joints on the hip of two front legs are adjusted from $[-30, 30]$ to $[-0.3, 0.3]$.
\begin{lstlisting}[language=Python]
# broken hip joints of front legs
<joint axis="0 0 1" name="hip_1" pos="0.0 0.0 0.0" range="-0.3 0.3" type="hinge"/>
<joint axis="0 0 1" name="hip_2" pos="0.0 0.0 0.0" range="-0.3 0.3" type="hinge"/>
\end{lstlisting}

\subsection{Adroit Dynamics Shifts}
For Adroit, we use Pen and Door. Each task is evaluated under two dynamics shift types, kinematic broken-joint and morphology shrink-finger, and two difficulty levels, medium and hard. These tasks require controlling a high-dimensional Shadow Hand and are substantially more challenging than locomotion tasks because of their high-dimensional action spaces and sparse rewards.

\textbf{\textit{Adroit-kinematic-broken-joint}}: We restrict the rotation ranges of selected joints in the index finger and the thumb. The medium level sets these ranges to one quarter of the source-domain ranges, while the hard level further reduces them to one eighth.
\begin{lstlisting}[language=Python]
# medium-level broken joint: index finger
<joint name="FFJ3" pos="0 0 0" axis="0 1 0" range="-0.109 0.109" user="1103"/>
<joint name="FFJ2" pos="0 0 0" axis="1 0 0" range="0 0.39275" user="1102"/>
<joint name="FFJ1" pos="0 0 0" axis="1 0 0" range="0 0.39275" user="1101"/>
<joint name="FFJ0" pos="0 0 0" axis="1 0 0" range="0 0.39275" user="1100"/>

# medium-level broken joint: thumb
<joint name="THJ4" pos="0 0 0" axis="0 0 -1" range="-0.26175 0.26175" user="1121"/>
<joint name="THJ3" pos="0 0 0" axis="1 0 0" range="0 0.32725" user="1120"/>
<joint name="THJ2" pos="0 0 0" axis="1 0 0" range="-0.0655 0.0655" user="1119"/>
<joint name="THJ1" pos="0 0 0" axis="0 1 0" range="-0.131 0.131" user="1118"/>
<joint name="THJ0" pos="0 0 0" axis="0 1 0" range="-0.3925 0" user="1117"/>
\end{lstlisting}

\textbf{\textit{Adroit-morphology-shrink-finger}}: We shorten the proximal, intermediate, and distal phalanges in the index, middle, ring, and little fingers. The medium level sets the phalange lengths to one quarter of their source-domain sizes, while the hard level sets them to one eighth. The thumb is kept unchanged.
\begin{lstlisting}[language=Python]
# medium-level shrink finger: index finger
<geom name="C_ffproximal" class="DC_Hand" size="0.01 0.005625" pos="0 0 0.005625" type="capsule"/>
<body name="ffmiddle" pos="0 0 0.01125">
<geom name="C_ffmiddle" class="DC_Hand" size="0.00805 0.003125" pos="0 0 0.003125" type="capsule"/>
<body name="ffdistal" pos="0 0 0.00625">
<geom name="C_ffdistal" class="DC_Hand" size="0.00705 0.003" pos="0 0 0.003" type="capsule" condim="4"/>
<site name="S_fftip" pos="0 0 0.0065" group="3"/>
\end{lstlisting}

\section{Algorithm Details}
\label{app:algorithm details}
We provide the pseudocode of TBM and TABB in Algorithm~\ref{alg:tbm} and Algorithm~\ref{alg:tabb} , respectively, and report the hyperparameter settings of TABB in Table~\ref{tab:hyp}.

\begin{algorithm}[t]
\caption{Target Bellman Mismatch Estimation}
\label{alg:tbm}
\begin{algorithmic}[1]
\State \textbf{Input:} Offline datasets $\mathcal{D}_{src}$ and $\mathcal{D}_{tar}$, number of representation learning steps $N_{rep}$, number of target refinement steps $N_{ref}$.
\State \textbf{Initialize:} State encoder $\phi$, state-action encoder $\psi$, target-domain predictor $f_{ref}$, anchor value function $V$.
\State Construct the mixed dataset $\mathcal{D}_{mix}=\mathcal{D}_{src}\cup\mathcal{D}_{tar}$.
\Statex \textit{// Shared latent representation learning}
\For{$i=1$ to $N_{rep}$}
    \State Sample a mini-batch from $\mathcal{D}_{mix}$.
    \State Obtain latent representations with Eq.~(\ref{eq:latent_rep}).
    \State Optimize $(\phi,\psi,f_{ref})$ with the representation loss in Eq.~(\ref{eq:tbm_loss}).
\EndFor
\Statex \textit{// Target-domain refinement}
\For{$i=1$ to $N_{ref}$}
    \State Sample a mini-batch $\{(s,a,r,s')\}$ from $\mathcal{D}_{mix}$.
    \State Freeze $(\phi,\psi)$ and refine $f_{ref}$ with Eq.~(\ref{eq:tbm}) on target-domain data.
\EndFor
\Statex \textit{// Anchor value learning}
\State Train the target-domain anchor value function $V$ on $\mathcal{D}_{tar}$ with frozen encoders.
\State \textbf{Return:} Encoders $\phi,\psi$, target-domain predictor $f_{ref}$, and anchor value function $V$.
\end{algorithmic}
\end{algorithm}

\begin{algorithm}[t]
\caption{Target-Aligned Bellman Backup (TABB)}
\label{alg:tabb}
\begin{algorithmic}[1]
\State \textbf{Input:} Offline datasets $\mathcal{D}_{src}$ and $\mathcal{D}_{tar}$, number of policy training steps $N_{\pi}$.
\State \textbf{Initialize:} Critic $Q_{\theta}$, value function $V_{\eta}$, policy $\pi_{\omega}$.
\State Obtain the encoders $\phi,\psi$, target-domain predictor $f_{ref}$, and anchor value function $V_{anc}$ by calling Algorithm~\ref{alg:tbm}.
\Statex \textit{// TBM-based source reweighting and policy learning}
\For{$j=1$ to $N_{\pi}$}
    \State Sample mini-batches $\{(s,a,r,s')\}$ from $\mathcal{D}_{src}$ and $\mathcal{D}_{tar}$.
    \State Compute the TBM of source transitions with Eq.~(\ref{eq:tbm}).
    \State Convert TBM into source transition weights with Eq.~(\ref{eq:weight}).
    \State Update the critic $Q_{\theta}$ with the TBM-weighted Bellman objective in Eq.~(\ref{eq:q_loss}).
    \State Update the value function $V_{\eta}$ on target-domain data with Eq.~(\ref{eq:v_loss}).
    \State Update the policy $\pi_{\omega}$ on target-domain data with Eq.~(\ref{eq:pi_loss}).
\EndFor
\State \textbf{Return:} Learned target-domain policy $\pi_{\omega}$.
\end{algorithmic}
\end{algorithm}

\begin{table}[h]
\centering
\caption{Hyperparameters of TABB.}
\label{tab:hyp}
\small
\renewcommand{\arraystretch}{1.1}
\setlength{\tabcolsep}{6pt}
\begin{tabular}{ll}
\toprule
\textbf{Hyperparameter} & \textbf{Value} \\
\midrule
\multicolumn{2}{l}{\textbf{Shared}} \\
\quad Actor network & $(256, 256)$ \\
\quad Critic network & $(256, 256)$ \\
\quad Learning rate & $3 \times 10^{-4}$ \\
\quad Optimizer & Adam \\
\quad Discount factor $\gamma$ & $0.99$ \\
\quad Target update rate & $5 \times 10^{-3}$ \\
\quad Batch size & $256$ \\
\quad Expectile parameter $\tau$ & $0.7$ \\
\quad Advantage temperature $\beta$ & $3.0$ \\
\midrule
\multicolumn{2}{l}{\textbf{Ours}} \\
\quad Latent state dimension & $64$ \\
\quad Latent action dimension & $64$ \\
\quad State encoder & $(256, 256)$ \\
\quad State-action encoder & $(256, 256)$ \\
\quad Latent pretraining steps & $2 \times 10^{5}$ \\
\quad Temperature $\tau_h$ & $0.3$ \\
\bottomrule
\end{tabular}
\end{table}

\section{The Use of Large Language Models (LLMs)}
We employ Large Language Models (LLMs) for grammar checking in our paper.

\section{Compute Resources}
Our method is trained with an cluster containing 30 nodes, each node is an Ubuntu 22.04 LTS server, with 4 $\times$ NVIDIA A40 GPU (Ampere architecture, 48GB VRAM), 72-core processor (dual-socket Intel Xeon Platinum), and 503GB memory.

\end{document}